\documentclass[10pt,twocolumn,letterpaper]{article}

\usepackage{cvpr}              % To produce the CAMERA-READY version
\usepackage{graphicx}
\usepackage{amsmath}
\usepackage{amssymb}
\usepackage{booktabs}

\usepackage{subcaption}
\usepackage{caption}
\usepackage[ruled]{algorithm2e}
\usepackage[table,xcdraw]{xcolor}
\usepackage{color}
\usepackage{xcolor}
\usepackage{enumitem}
\usepackage[numbers]{natbib}
\usepackage{float}
\usepackage{bm}
\usepackage{pifont}
\usepackage{gensymb}

\usepackage{multirow}
\usepackage[table]{xcolor}
\usepackage{arydshln}
\usepackage{threeparttable}
\usepackage{colortbl}
\usepackage{bm}
\usepackage{bbm}
\usepackage{array}
\usepackage{siunitx}
\usepackage{url}

\usepackage{mathrsfs}

\newcommand{\ourmodel}{\textsc{IS-Fusion}}
\makeatletter
\newcommand{\thickhline}{%
    \noalign {\ifnum 0=`}\fi \hrule height 1pt
    \futurelet \reserved@a \@xhline
}
\makeatother
\usepackage[pagebackref,breaklinks,colorlinks]{hyperref}

\def\cvprPaperID{3596} % *** Enter the CVPR Paper ID here
\def\confName{CVPR}
\def\confYear{2024}

\begin{document}

\title{\!\!\!\!\textsc{IS-Fusion}: Instance-Scene Collaborative Fusion for Multimodal 3D Object Detection\!\!\!\!\!\!\!\!\!\!\!}

%% omit the following lines up until the closing ``}''.
%% Additional authors and addresses can be added with ``\and'',
%% just like the second author.
%% To save space, use either the email address or home page, not both
%\and
%{\tt\small secondauthor@i2.org}-
%}

\author{Junbo Yin\textsuperscript{1}, Jianbing Shen\textsuperscript{2}, Runnan Chen\textsuperscript{3}, Wei Li\textsuperscript{4*}, Ruigang Yang\textsuperscript{4},
 Pascal Frossard\textsuperscript{5}, Wenguan Wang$^{6}\thanks{Corresponding author.}$~\!\\
\small{\quad$^1$}\small School of Computer Science and Technology, Beijing Institute of Technology \quad \small{$^2$}\small SKL-IOTSC, CIS, University of Macau  \\
\!\! \small{$^3$}\small The University of Hong Kong\quad \small{$^4$}Inceptio\quad
\small{$^5$}\small  \'{E}cole Polytechnique F\'{e}d\'{e}rale de Lausanne (EPFL) \quad
\small{$^6$}\small ReLER, CCAI, Zhejiang University\\
% Beijing Institute of Technology\\
{\tt\small \{yinjunbocn,wenguanwang.ai\}@gmail.com}
}

%\footnotetext[2]{Corresponding author: author2@example.com}

\maketitle

%%%%%%%%% ABSTRACT
\begin{abstract}

Bird's eye view (BEV) representation has emerged as a dominant solution for describing 3D space in autonomous driving scenarios. However, objects in the BEV representation typically exhibit small sizes, and the associated point cloud context is inherently sparse, which leads to great challenges for reliable 3D perception. 
{In this paper, we propose \textbf{\ourmodel}, an innovative multimodal \textbf{fusion} framework that jointly captures the \textbf{I}nstance- and \textbf{S}cene-level contextual information. \ourmodel~essentially differs from existing approaches that only focus on the BEV scene-level fusion by explicitly incorporating instance-level multimodal information, thus facilitating the instance-centric tasks like 3D object detection.}
It comprises a Hierarchical Scene Fusion (HSF) module and an Instance-Guided Fusion (IGF) module. 
%HSF applies Point-to-Grid and Grid-to-Region transformers to capture the multimodal scene context at different granularities. This leads to enriched scene-level features that are later used to generate high-quality instance features. Next, IGF explores the relationships between these instances, and aggregates the local multimodal context for each instance. These instances then serve as guidance to enhance the scene feature and yield an instance-aware BEV representation. On the challenging nuScenes benchmark, \ourmodel~outperforms all the published multimodal works to date. Code is available at: \url{https://github.com/yinjunbo/IS-Fusion}.
HSF applies Point-to-Grid and Grid-to-Region transformers to capture the multimodal scene context at different granularities. IGF mines instance candidates, explores their relationships, and aggregates the local multimodal context for each instance. These instances then serve as guidance to enhance the scene feature and yield an instance-aware BEV representation. On the challenging nuScenes benchmark, \ourmodel~outperforms all the published multimodal works to date. Code is available at: \url{https://github.com/yinjunbo/IS-Fusion}.
%Notably, it achieves 72.8\% mAP on the nuScenes validation set, outperforming prior art like BEVFusion by 4.3\% mAP. 
%Extensive experiments on the challenging nuScenes benchmark demonstrate that \ourmodel~outperforms all the published works to date. Our work provides new perspectives to current BEV models by prioritizing crucial instance-level information.
% 

\end{abstract}

%%%%%%%%% BODY TEXT

\section{Introduction}
\label{introduction}

3D object detection~\cite{qi2018frustum,zhu2023curricular,he2023msf,zhou2023octr,yin2021graph,meng2021towards} is a critical task in various applications such as autonomous driving and robotics. 
Over the past few years, tremendous progress has been achieved in point cloud-based 3D object detection, due to the effective 3D neural network models~\cite{qi2017pointnet++,gilmer2017neural,graham20183d,zhao2021point,feng2023clustering}. While point clouds, typically captured by depth-aware sensors such as LiDAR, provide valuable geometric information about the 3D space, {they often lack detailed texture descriptions and are sparsely distributed over long distances, \eg, beyond 100 meters in outdoor scenarios like nuScenes~\cite{caesar2020nuscenes}.} {To tackle these limitations, a recent trend is to perform multimodal 3D object detection~\cite{liu2022bevfusion,liang2022bevfusion,li2022unifying,yang2022deepinteraction,li2023lwsis} by fusing information from both point clouds and synchronized multi-view images. The image modality provides detailed texture and dense semantic information~\cite{wu2023virtual,han2022brnet}, which complements the sparse point cloud and thus enhances the 3D perception capacity.}

%----------- Figure 1 ----------- %
\begin{figure}
\includegraphics[width=0.99\linewidth]{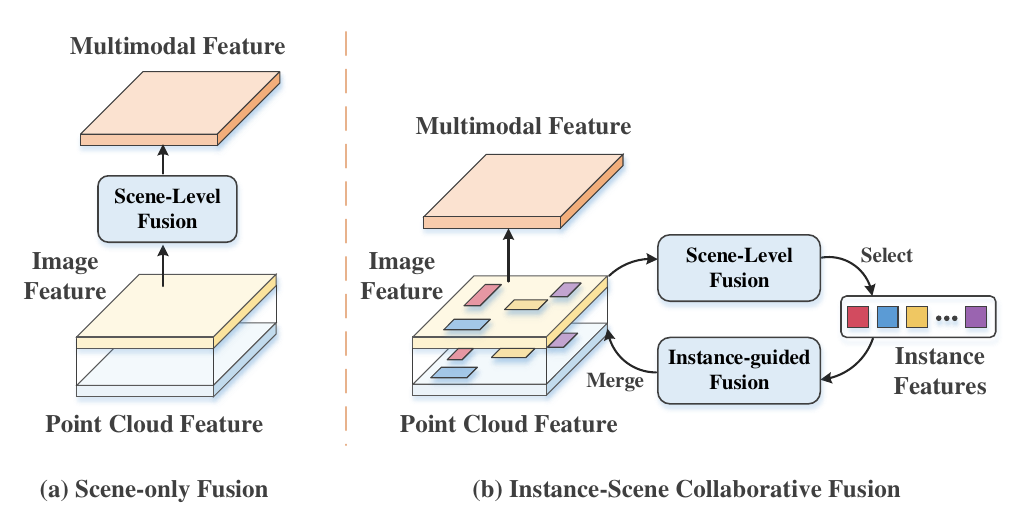}
\centering
 \vspace{-2mm}
 \caption{ \textbf{Motivation of \ourmodel.} (a) Previous approaches typically focus on fusion at the entire scene level during multimodal encoding. (b) In contrast, \ourmodel~places additional emphasis on the fusion at the instance level and explores the instance-to-scene collaboration to enhance the overall representation.}
\label{fig:motivation}
 \vspace{-2mm}
\end{figure}

To handle heterogeneous data from different modalities, existing approaches~\cite{liang2022bevfusion,liu2022bevfusion,liu2023bird} typically pre-define a unified space that is compatible with both modalities (\ie, the bird's eye view (BEV) in the ego-vehicle coordinate system), and then perform feature alignment and fusion on this shared space. {BEV representation simplifies the complex 3D space into a 2D plane, making it easier to understand the scene.} However, performing fusion from the entire BEV scene level ignores the inherent difference between the foreground instances and background regions, which may undermine the performance. For example, object instances represented in the BEV often exhibit smaller sizes compared to those observed in natural images. Additionally,  the number of BEV grid cells occupied by foreground instances is significantly lower than those occupied by background ones, leading to a severe imbalance between foreground and background samples. As a result, the above approaches struggle to capture local context around the object instances, or largely rely on additional networks in the decoding stage to iteratively refine the detections~\cite{yang2022deepinteraction,bai2022transfusion}.
{While a few methods~\cite{li2023fully,cai2023objectfusion} aim to perform object-level encoding, they ignore the potential collaboration between the scene and instance features. For example, a false negative object in a scene can be potentially rectified by enhancing its feature through interactions with the instances sharing similar semantic information.}
Therefore, it remains an open question \textit{how to simultaneously formulate the instance-level and scene-level context, as well as elegantly integrate them by leveraging multimodal fusion.}

In this work, we present a new multimodal detection framework, \ourmodel, to tackle the above challenge. As shown in Fig.~\ref{fig:motivation}, \ourmodel~explores both the \textbf{I}nstance-level and \textbf{S}cene-level \textbf{Fusion}, as well as encourages the interaction between the instance and scene features to strengthen the overall representation.
It consists of two crucial components: the Hierarchical Scene Fusion (HSF) module and the Instance-Guided Fusion (IGF) module. HSF aims to capture scene features at various granularities by utilizing Point-to-Grid and Grid-to-Region transformers. This also enables the generation of high-quality instance-level features that are vital for IGF. In IGF, {the foreground instance candidates are determined by the heatmap scores of the scene feature}; meanwhile, an inter-instance self-attention is employed to capture the instance relationships. These instances then aggregate essential semantic information from the multimodal context through deformable attention. Furthermore, we incorporate an Instance-to-Scene transformer attention to enforce the local instance features to collaborate with the global scene feature. This yields an enhanced BEV representation that is better suited for instance-aware tasks like 3D object detection.

In summary, \ourmodel~provides a new insight into existing multimodal 3D detection approaches that focus on scene-level fusion. By incorporating HSF and IGF, it explicitly promotes collaboration between scene- and instance-level features, thereby ensuring comprehensive representation and yielding improved detection results. 
  Extensive experiments on the competitive nuScenes~\cite{caesar2020nuscenes} dataset demonstrate that \ourmodel~attains the best performance among all the published 3D object detection works. For example, it achieves 72.8\% mAP on the nuScenes validation set, outperforming prior art BEVFusion~\cite{liu2022bevfusion} by 4.3\% mAP. It also surpasses concurrent works like CMT~\cite{yan2023cross} and SparseFusion~\cite{li2023fully} by 2.5\% and 1.8\% mAP, respectively. 

%----------- Figure 2 ----------- %
\begin{figure*}
\includegraphics[width=0.99\linewidth]{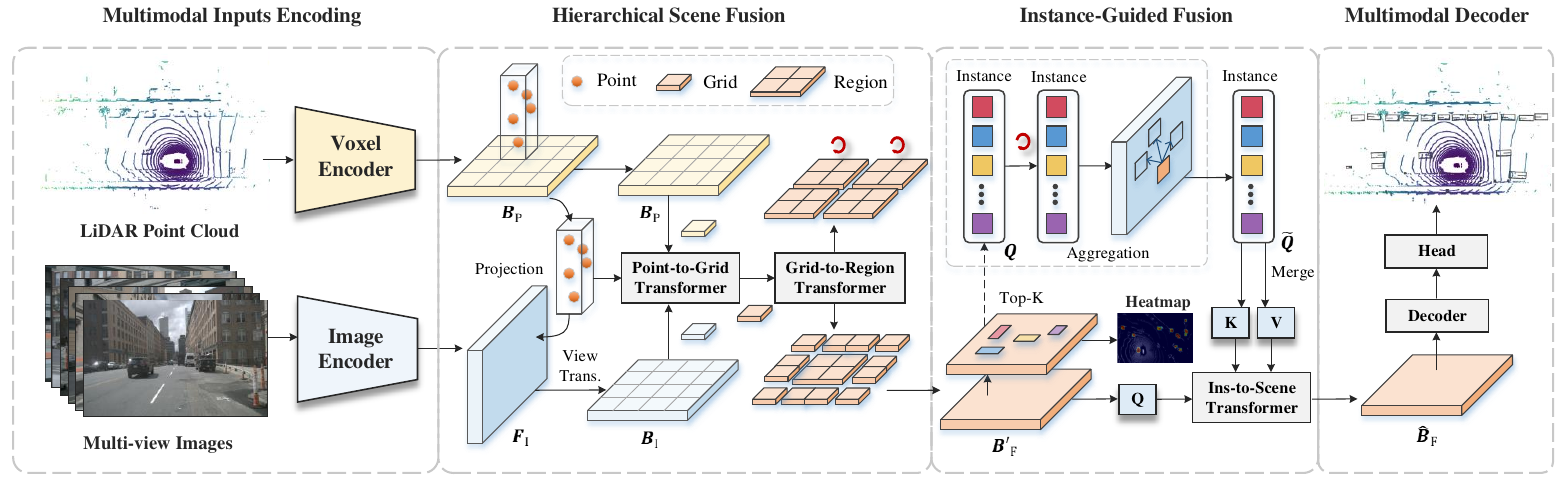}
\centering
 % \vspace{-2mm}
 \caption{ \textbf{Overview of our~\ourmodel~framework.} Multimodal inputs including a point cloud and multi-view images are first processed by modality-specific encoders to obtain initial features. Then, the HSF module, equipped with Point-to-Grid and Grid-to-Region transformers, utilizes these features to generate a scene-level feature with hierarchical context. Furthermore, the IGF module identifies the top-$K$ salient instances and aggregates the multimodal context for each instance. Finally, these instances are employed by the Instance-to-Scene transformer to propagate valuable information to the scene, producing the final BEV representation with improved instance awareness.}
%Finally, an Instance-to-Scene transformer employs the scene-level queries ($Q$) to attend the instance-level keys ($K$) and values ($V$), producing the final BEV representation with improved instance awareness. 
\label{fig:pipeline}
%\vspace{-2mm}
\end{figure*}
%Then, the points in each BEV grid (\ie, a pillar) are used as proxy to fetch coordinate-corresponding image features.

\section{Related Work}
\label{related_work}

\noindent\textbf{LiDAR-based 3D Object Detection.} LiDAR sensors are essential for advanced autonomous driving due to their capacity of perceiving objects in 3D space even in adverse illumination and weather environments, where they are usually more reliable than camera sensors~\cite{li2022bevformer,liu2022petr,wang2019pseudo,li2019stereo,feng2024towards,wang2021depth}. Current LiDAR-based detection approaches can be broadly classified into three categories according to the various encoding formats of point cloud: point-based~\cite{shi2019pointrcnn,yang20203dssd,qi2019deep,zhou2020joint,shi2020point,feng2024interp,yin2022proposalcontrast}, voxel-based~\cite{zhou2018voxelnet,lang2019pointpillars,he2022voxel,xiang2023di,wang2023ssda3d,yin2022semi} and point-voxel fusion networks~\cite{shi2020pv,chen2019fast,yang2019std,miao2021pvgnet,hu2022point}. Shi~\textit{et al.}\cite{shi2019pointrcnn} propose an early work for point-wise 3D detection by extending PointNet~\cite{qi2017pointnet,qi2017pointnet++} backbone with a two-stage proposal refinement network. Due to the huge computation overhead in large-scale scenes with more than 100k points, point downsampling operations~\cite{yang20203dssd} have to be applied. A more popular solution is to use the voxel-based representation, where the point clouds are quantified by regular grids such that standard convolutional networks can be directly applied. VoxelNet~\cite{zhou2018voxelnet} is the seminal work that exploits the 3D convolutional network that is later optimized in~\cite{yan2018second} with sparse convolution. In addition, there are also some works like~\cite{shi2020pv} exploring joint point-voxel representation by enhancing the region proposals with the raw points information, while they often require multiple stages to refine the 3D proposals.

These LiDAR-only 3D detectors usually operate with sparse and noisy context provided by point cloud data. However, in challenging scenarios where objects have low reflectivity, small sizes or are heavily occluded, relying solely on point cloud data may lead to inaccurate detection. Therefore, our focus is to explore the multimodal context by incorporating the merits of both geometry-aware point clouds and semantic-rich images to guarantee advanced 3D object detection capability.

\noindent\textbf{LiDAR-camera Fusion for 3D Object Detection.} Multimodal 3D object detection~\cite{chen2017multi,liang2018deep,liang2019multi,xu2018pointfusion} has recently received considerable attention. It also has been proven that multimodal learning can yield a more accurate latent space representation~\cite{huang2021makes} compared to the unimodal learning. Multimodal fusion approaches for 3D object detection basically comprise early fusion~\cite{vora2020pointpainting,vora2020pointpainting,chen2022autoalignv2,xu2021fusionpainting,yin2021multimodal}, middle fusion~\cite{li2022deepfusion,li2022voxel,liu2022bevfusion,liang2022bevfusion,li2022unifying,yang2022deepinteraction,zhang2022cat,piergiovanni20214d} and late fusion~\cite{chen2017multi,yang2022graph,bai2022transfusion,li2023logonet,liang2019multi,ku2018joint}, which are categorized based on the stages at which data fusion occurs. 

{The works in~\cite{vora2020pointpainting, wang2021pointaugmenting} are pioneering efforts of early fusion that enhance input points with corresponding image pixel features. Later, Chen~\textit{et al.}~\cite{chen2022autoalignv2} propose to fuse point cloud and image features at the voxel level and aggregate information from multiple sampling points. However, the early fusion approaches are more sensitive to potential calibration errors. Late fusion approaches like \cite{chen2017multi,ku2018joint} typically fuse multimodal information at the region proposal level, where the region proposals are usually generated separately by modality-specific encoders. These approaches may result in limited interactions between modalities during proposal generation~\cite{li2023fully}, leading to suboptimal detection performance. 
By contrast, middle fusion has become increasingly popular recently as it encourages multimodal feature interaction at various representation stages, which is more robust to calibration error.
%The state-of-the-art methods includes BEVFusion~\cite{liu2022bevfusion,liang2022bevfusion} and DeepInteraction~\cite{yang2022deepinteraction}. 
Liu~\textit{et al.}~\cite{liu2022bevfusion} and Liang~\textit{et al.}~\cite{liang2022bevfusion} propose to align point cloud with multi-view image features on a unified BEV plane to simplify the fusion process. Yang~\textit{et al.}~\cite{yang2022deepinteraction} further suggest fusion from both BEV and image perspective-view spaces to preserve modality-specific information.}

{Unlike these approaches \cite{liu2022bevfusion, yang2022deepinteraction,yan2023cross}, which primarily integrate point cloud and image features at a global scene level, our~\ourmodel~investigates fusion from both local instance level and global scene level. This permits the advantages of  `hybrid fusion', which also marks an improvement over some concurrent works~\cite{li2023fully,cai2023objectfusion} that focus only on the instances and ignore the collaboration with the scene. \ourmodel~smartly exchanges information between instances and the scene, and thus facilitates the instance-centric tasks such as 3D object detection, as shown later. }

\section{Methodology}
\label{method}

We first introduce the general overview of the proposed \ourmodel~in Sec.~\ref{sec:framework}. Next, we delve into the details of the HSF module in Sec.~\ref{sec:HSF}. After that, in Sec.~\ref{sec:IGF}, we elaborate on the crucial design steps of the IGF module. 

\subsection{Overall Framework}
\label{sec:framework}
As illustrated in Fig.~\ref{fig:pipeline},  each scene is represented by a LiDAR point cloud $P$, along with synchronized RGB images $I\!=\!\{{I}_1, {I}_2, \ldots, {I}_N\}$ captured by $N$ cameras that are well-calibrated with the LiDAR sensor. Our goal is to devise a detection model capable of producing precise 3D bounding boxes $\bm{Y}$, given multimodal inputs $(P, I)$. Formally, the proposed \ourmodel~model is defined by:
\vspace{-1mm}
\begin{equation}
\bm{Y}={f}_\text{dec}({f}_\text{enc}(f_{\text{point}}(P), f_\text{img}(I))),
\end{equation}	
where $f_\text{point}(\cdot)$ and $f_\text{img}(\cdot)$ serve as the input encoding modules, ${f}_\text{enc}(\cdot)$ denotes the multimodal encoder (formed by HSF and IGF) and ${f}_\text{dec}(\cdot)$ is the decoder.

\noindent\textbf{Multimodal Input Encoding.} 
To handle inputs from heterogeneous modalities, we first utilize modality-specific encoders to get their respective initial representation, \ie, $\bm{B}_\text{P}\!=\!f_\text{point}(P)$ and $\bm{F}_\text{I}\!=\!f_\text{img}(I)$. Following~\cite{liu2022bevfusion,yang2022deepinteraction}, we instantiate $f_\text{point}(\cdot)$ with VoxelNet~\cite{zhou2018voxelnet}, and $f_\text{img}(\cdot)$ by Swin-Transformer~\cite{liu2021swin}.
This yields the point cloud BEV feature $\bm{B}_\text{P}$ and the image Perspective-View (PV) features $\bm{F}_\text{I}$. 
In particular, $\bm{B}_{\text{P}}\!\in\!\mathbb{R}^{W\!\times\!{H}\!\times\!{C}\!}$ is obtained by compressing the height dimension of the 3D voxel feature as in~\cite{zhou2018voxelnet}, where $W$ and $H$ are the numbers of BEV grid cells along the $x$ and $y$ axes, and $C$ denotes the channel dimension.

\noindent\textbf{Multimodal Encoder.}  The multimodal encoder ${f}_\text{enc}(\cdot)$ conducts cross-modality feature fusion between $\bm{B}_\text{P}$ and $\bm{F}_\text{I}$ to yield a fused BEV feature $\hat{\bm{B}}_\text{F}\!\in\!\mathbb{R}^{W\!\times\!{H}\!\times{C}}$. {In contrast to previous multimodal encoders that only focus on fusion at the entire scene level~\cite{liu2022bevfusion,yang2022deepinteraction}, we develop both instance-level and scene-level representations.} To this end, we design ${f}_\text{enc}(\cdot)$ using two modules, namely, HSF module ${f}_\text{HSF}(\cdot)$ and IGF module ${f}_\text{IGF}(\cdot)$:
\begin{equation}
\hat{\bm{B}}_\text{F}={f}_\text{enc}(\bm{B}_\text{P}, \bm{F}_\text{I})={f}_\text{IGF}({f}_\text{HSF}( \bm{B}_\text{P}, \bm{F}_\text{I})),
\label{eq:overall}
\end{equation}
where ${f}_\text{HSF}(\cdot)$  generates multi-granularity scene feature, while ${f}_\text{IGF}(\cdot)$ further integrates crucial information about foreground instances. We will elaborate on ${f}_\text{HSF}(\cdot)$ and ${f}_\text{IGF}(\cdot)$ in Sec.~\ref{sec:HSF} and Sec.~\ref{sec:IGF}, respectively. 

\noindent\textbf{Multimodal Decoder. } The multimodal decoder aims to yield the final 3D detections $\bm{Y}$ based on the BEV representation $\hat{\bm{B}}_\text{F}$, given by $\bm{Y}\!=\!{f}_\text{dec}(\hat{\bm{B}}_\text{F})$. In our work, $ {f}_\text{dec}(\cdot)$ is built upon a transformer architecture~\cite{yao2021efficient}, which contains several attention layers and a feed-forward-network serving as the detection head. 
During training, the Hungarian algorithm~\cite{kuhn1955hungarian} is applied for matching the predicted and ground-truth bounding boxes. Meanwhile, Focal loss~\cite{lin2017focal} and $L$1 loss are adopted for the classification and 3D bounding box regression, respectively.

\subsection{Hierarchical Scene Fusion}
\label{sec:HSF} 
\begin{figure}
\includegraphics[width=0.95\linewidth]{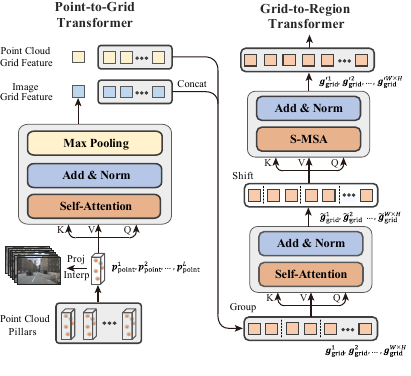}
\centering
 % \vspace{-4mm}
 \caption{ \textbf{Illustration of HSF module.} It first aggregates the point-level features into the grid-level features with the Point-to-Grid transformer, and then explores the inter-grid and inter-region feature interaction through the Grid-to-Region transformer.}
\label{fig:HSF}
 % \vspace{-2mm}
\end{figure}

Given the point cloud BEV feature $\bm{B}_\text{P}$ and the image PV feature $\bm{F}_\text{I}$, we suggest a Hierarchical Scene Fusion (HSF) module $f_\text{HSF}(\cdot)$ to integrate $\bm{B}_\text{P}$ and $\bm{F}_\text{I}$ and obtain the fused scene representation $\bm{B}_\text{F}\!\in\!\mathbb{R}^{W\!\times\!{H}\!\times\!{C}}$. To be specific, $f_\text{HSF}(\cdot)$ comprises a Point-to-Grid transformer $f_\text{P2G}(\cdot)$ and Grid-to-Region transformer $f_\text{G2R}(\cdot)$, given by:
\begin{equation}
\bm{B}_\text{F}=f_\text{HSF}(\bm{F}_\text{I},  \bm{B}_\text{P})=f_\text{G2R}(f_\text{P2G}(\bm{F}_\text{I}),  \bm{B}_\text{P}).
\label{eq:hsf}
\end{equation}
Here, $f_\text{P2G}(\cdot)$ considers the inter-point/pixel correlations in each BEV grid, while $f_\text{G2R}(\cdot)$ further mines the inter-grid and inter-region multimodal scene context. The intuition is that different feature granularities capture scene context at different levels. For example, at the point level, each element provides detailed information about specific components of an object. In contrast, features at the grid/region level are capable of capturing the broader scene structure and the distribution of objects. HSF fully leverages various representation granularities, as illustrated in Fig.~\ref{fig:HSF}.

%----------- Figure 3 ----------- %

\noindent\textbf{Point-to-Grid Transformer.} 
Let us denote $G\!=\!\{{g}_\text{grid}^{1},$ $ {g}_\text{grid}^2, \ldots, {g}_\text{grid}^{W\times{H}}\}$ as the BEV grid cells obtained by discretizing the point cloud scene $P$ into pillars following~\cite{lang2019pointpillars}. Each grid cell $g_\text{grid}\!\in\!{G}$ is a pillar containing $L$ points $\{{p}_\text{point}^{1},$ $ {p}_\text{point}^2, \ldots, {p}_\text{point}^L\}$. The Point-to-Grid transformer assigns each point with its corresponding image feature and aggregates them into a BEV grid-wise feature. 

%Speciﬁcally, we first discretize the point cloud scene $P$ into BEV grids $\{g\}$ following~\cite{lang2019pointpillars}, where each grid $g_{i}\in\{{g}_1, {g}_2, \ldots, {g}_{W\times{H}}\}$ is a pillar containing $L$ points $\{{p}_1, {p}_2, \ldots, {p}_L\}$. 
Specifically, we project the $L$ points within a pillar $[{p}_\text{point}^{1}, {p}_\text{point}^2, \ldots, {p}_\text{point}^L]\!\in\!\mathbb{R}^{L\times{3}}$ onto the image feature map $\bm{F}_\text{I}$ and retrieve their pixel-level features:
\vspace{-1mm}
\begin{equation}
\begin{aligned}
&\!\!\!\![u^1, u^2, \cdots, u^{L}] \!=\! f_\text{proj}([{p}_\text{point}^{1}, {p}_\text{point}^2, \ldots, {p}_\text{point}^L]),  \\
&\!\!\!\![\bm{p}_\text{point}^{1}, \bm{p}_\text{point}^2, \ldots, \bm{p}_\text{point}^L] \!=\! f_\text{interp}(\bm{F}_\text{I}, [u^1, u^2, \cdots, u^{L}]),
\label{eq:proj}
\end{aligned}
\end{equation}
where $f_\text{proj}(\cdot)$ indicates the projection process from point cloud to multi-view images that yields 2D coordinates $[u^1, u^2, \cdots, u^{L}]$ on the image plane, and  $f_\text{interp}(\cdot)$ is the bilinear interpolation function computing features at non-integer coordinates. In this way, we get the point-wise features $[\bm{p}_\text{point}^{1}, \bm{p}_\text{point}^2, \ldots, \bm{p}_\text{point}^L]\in\mathbb{R}^{L\times{C}}$.

To handle the potential calibration noise between LiDAR and cameras, our Point-to-Grid transformer compares all the points within a pillar. This enables each point to consider a larger receptive field and implicitly rectifies noisy points. Afterwards, we merge the point-wise information with a max pooling operation $f_\text{max}(\cdot)$:  
\begin{equation}
%\begin{aligned}
\!\!\!\!\bm{g}_\text{grid}\!=\!f_\text{max}({f}_\text{MSA}([\bm{p}_\text{point}^{1}, \bm{p}_\text{point}^2, \ldots, \bm{p}_\text{point}^L]))\!\in\!\mathbb{R}^{1\times C},
%& \bm{g}_{I} = \max_{l}(\bm{p}'_{l}) \in \mathbb{R}^{1 \times C},
%\end{aligned}
\end{equation}
where ${f}_\text{MSA}(\cdot)$ is the multi-head self-attention~\cite{vaswani2017attention}, and $\bm{g}_\text{grid}$ is a grid-wise feature that will be assigned to the image BEV feature $\bm{B}_\text{I}\!\in\!\mathbb{R}^{W\times{H}\times C}$.
 Then, we compute the multimodal BEV feature $\bm{B}_\text{F}$ by combing $\bm{B}_\text{I}$ with the point cloud BEV feature $\bm{B}_\text{P}\!\in\!\mathbb{R}^{W\times{H}\times C}$:
 \vspace{-2mm}
\begin{equation}
	\bm{B}_\text{F} = f_\text{conv}([\bm{B}_\text{I}, \bm{B}_\text{P}])\in\mathbb{R}^{W\times{H}\times C},
	\label{eq:bevfusion}
\end{equation}
where $[\cdot, \cdot]$ denotes the concatenation, and $f_\text{conv}(\cdot)$ is implemented by a 3$\times$3 convolutional layer.
% where a convolutional layer is applied on the concatenation of the point cloud and image grid features, and $\bm{g}$ is an element of the multimodal BEV scene feature $\bm{G}\in\mathbb{R}^{W\times{H}\times C}$. 
 
\noindent\textbf{Grid-to-Region Transformer.} In addition to the Point-to-Grid transformer that models the inter-point dependencies, we further explore the inter-grid and inter-region relationships via the Grid-to-Region transformer to capture the global scene context. This can be denoted as $\bm{B}^{\prime}_\text{F}\!=\!{f}_\text{G2R}(\bm{B}_\text{F})$, where $\bm{B}^{\prime}_\text{F}$ is the enhanced BEV feature.

Intuitively, ${f}_\text{G2R}(\cdot)$ can be achieved by applying global self-attention to all the grid-wise features $\{\bm{g}_\text{grid}^{1},\bm{g}_\text{grid}^2,  \cdots, $ $\bm{g}_\text{grid}^{W\times{H}}\}\!\in\!\bm{B}_\text{F}$. However, this can be computationally expensive due to the large number of grid cells. Hence, we choose to group these grid features into different regions following~\cite{dosovitskiy2020image}. Each region is a subset described by $M\!\times\!{M}$ grid cells $\{{g}_\text{grid}^{1}, {g}_\text{grid}^2, \cdots, {g}_\text{grid}^{M^{2}}\}$. Next, we view each region as a whole and exchange information between the grids in a region through inter-grid attention. 
More concretely, this is realized by the multi-head attention ${f}_\text{MSA}(\cdot)$ operating on a set of grid-wise features $[\bm{g}_\text{grid}^{1}, \bm{g}_\text{grid}^2, \cdots, \bm{g}_\text{grid}^{M^{2}}]\in\mathbb{R}^{M^{2}\times{C}}$:
%\begin{equation}
%\bm{B'}_\text{F}={f}_\text{G2R}(\{\bm{g}_1, \bm{g}_2, \ldots, \bm{g}_{W\times{H}}\})\in\mathbb{R}^{{W\times{H}}\times{C}}.
%\label{eq:g2r}
%\end{equation}
%Intuitively, Since directly implementing ${f}_\text{G2R}(\cdot)$ as a global self-attention operation is computationally expensive, we choose to group the grid features into different regions following~\cite{dosovitskiy2020image}. In particular, each region is described by $M\times{M}$ grid cells $\{\bm{g}_1, \bm{g}_2, \cdots, \bm{g}_{M^{2}}\}$. Next, we view each region as a whole and exchange information between the grids in a region through inter-grid attention. Meanwhile, we capture the interactions between different regions with inter-region attention to model the global context. 
%%Since directly performing global attention on all the grids of the multimodal BEV feature $\bm{G}$ could be computationally expensive (\eg, involving 180$\times$180 queries), we choose to group $\bm{G}$ into different regions following~\cite{dosovitskiy2020image}, where each region corresponds an area containing $M\times{M}$ BEV grids $\{{g}_1, {g}_2, \cdots, {g}_{M^{2}}\}$. 
%
%%Speciﬁcally, let $\bm{g}\in\mathbb{R}^{M^{2}\times{C}}$ present a group of grid features from the same region. The inter-grid attention is computed as:
%More concretely, the inter-grid attention is computed by the self-attention operator ${f}_\text{MSA}(\cdot)$:
%\begin{equation}
%\begin{aligned}
%&	 \bm{g}' = \text{LN}(\text{MSA}(\bm{g})+\bm{g}), \\
%&	 \hat{\bm{g}} = \text{LN}(\text{MLP}(\bm{g}')+\bm{g}),
%\end{aligned}
%\end{equation}
\vspace{-1mm}
\begin{equation}
\!\!\!\![\tilde{\bm{{g}}}_\text{grid}^{1}, {\tilde{\bm{{g}}}}_\text{grid}^2, \cdots, \tilde{\bm{{g}}}_\text{grid}^{M^{2}}]\! = \!{f}_\text{MSA}([\bm{g}_\text{grid}^{1}, \bm{g}_\text{grid}^2, \cdots, \bm{g}_\text{grid}^{M^{2}}]), 
\end{equation}
where $[{\tilde{\bm{{g}}}}_\text{grid}^{1}, \tilde{\bm{{g}}}_\text{grid}^2, \cdots, \tilde{\bm{{g}}}_\text{grid}^{M^{2}}]$ are the attentive grid cells. 

Then, we capture the interactions between different regions with inter-region attention. To this end, we shift each region by $(M/2, M/2)$ grid cells and conduct self-attention on each shifted region containing $M\times{M}$ grid-wise features (using padding if necessary). This is given by:
%The inter-grid attention models the local relationships and dependencies among grid cells within the same region. Next, the inter-region attention enforces each region to exchange information with its neighbor regions. 
% such that each gird in the shifted region can attend grids belong to other region thorough self-attention, which is:
%through shifting each region by (M/2, M/2) grids in the BEV space. In this case, each region can receive knowledge from its neighbor regions. This can be realized by a shifted-window attention in~\cite{liu2021swin}: 
%\begin{equation}
%\begin{aligned}
%&	 \hat{\bm{g}}' = \text{LN}(\text{S-MSA}(\hat{\bm{g}})+\hat{\bm{g}}), \\
%&   \tilde{\bm{g}} = \text{LN}(\text{MLP}( \hat{\bm{g}}')+ \hat{\bm{g}}).
%&	 \end{aligned}
%\end{equation}
\vspace{-1mm}
\begin{equation}
[{\bm{g}^{\prime\triangle_1}_\text{grid}}, {\bm{g}^{\prime\triangle_2}_\text{grid}}, \cdots, {\bm{g}^{\prime\triangle_{M^{2}}}_\text{grid}}] \!=\! {f}_\text{S-MSA}([{\tilde{\bm{{g}}}}_\text{grid}^{\triangle_1}, \tilde{\bm{{g}}}_\text{grid}^{\triangle_2}, \cdots, \tilde{\bm{{g}}}_\text{grid}^{\triangle_{M^{2}}}]), 
\end{equation}
where ${f}_\text{S-MSA}(\cdot)$ indicates the shifted-window self-attention in~\cite{liu2021swin}, and $\{\triangle_{1}, \triangle_{2}, \cdots, \triangle_{M^{2}}\}$ represent the new grid indices after the shift. This allows each grid to interact with the grids coming from various regions before the shift, thus capturing long-range dependencies. Then, we rearrange all the attentive grid features $\{{\bm{g}^{\prime1}_\text{grid}}, {\bm{g}^{\prime2}_\text{grid}}, \cdots, {\bm{g}^{\prime W\times{H}}_\text{grid}}\}$ to get the enriched BEV feature map $\bm{B}^{\prime}_\text{F}\in\mathbb{R}^{W\times{H}\times{C}}$.

By exploiting the hierarchical representation, HSF enables the propagation of information from individual points to different BEV regions. This facilitates the integration of both local and global multimodal scene contexts. 
%The improved scene feature $\bm{H}$ is then be utilized in IGF to extract the instance-level context.

\subsection{Instance-Guided Fusion}
\label{sec:IGF}
\begin{figure}
\includegraphics[width=0.99\linewidth]{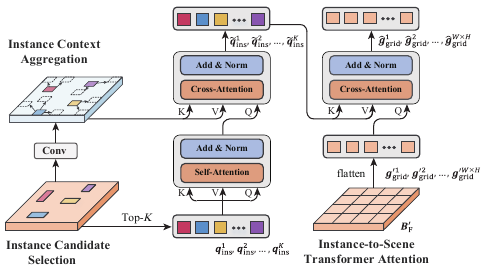}
\centering
% \vspace{-2mm}
 \caption{ \textbf{Illustration of IGF module.} Instance candidates are first initialized based on the BVE heatmap. Then, we perform reasoning on these instances, meanwhile aggregating rich semantic context from the image features. Finally, these instances transfer contextual information to the BEV scene feature through an Instance-to-Scene transformer attention mechanism.}
\label{fig:IGF}
  % \vspace{-2mm}
\end{figure}

The basic idea of IGF is to mine the multimodal context around each object instance (\eg, the lanes beside the vehicles), meanwhile integrating essential instance-level information into the scene feature. For example, if an object is incorrectly categorized as part of the background in the scene feature, we can rectify this by comparing it with all the relevant instances. Formally, given the scene feature $\bm{B}^{\prime}_\text{F}$ produced by HSF, ${f}_\text{IGF}(\cdot)$ in Eq.~(\ref{eq:overall}) is formulated as:
%\begin{equation}
%\hat{\bm{H}} = \mathcal{I}_{inst}(\bm{H})=\mathcal{I}_{i2s}(\bm{H}, \mathcal{I}_{agg}(\mathcal{I}_{sel}(\bm{H})))\in\mathbb{R}^{W\times{H}\times{C}},
%\label{eq:ins}
%\end{equation}
%\begin{align}
%& \{\bm{q}_0, \bm{q}_1, \ldots, \bm{q}_K\}={f}_\text{select}(\bm{B'}_\text{F})\in\mathbb{R}^{K\times{C}}, \\
%&\{{\bm{\tilde{q}}}_0,  {\bm{\tilde{q}}}_1, \ldots,  {\bm{\tilde{q}}}_K\}={f}_\text{agg}(\{\bm{q}_0, \bm{q}_1, \ldots, \bm{q}_K\}, \bm{B'}_\text{F}), \\
%& \bm{\hat{B}}_\text{F}={f}_\text{I2S}(\bm{B'}_\text{F}, \{{\bm{\tilde{q}}}_0,  {\bm{\tilde{q}}}_1, \ldots,  {\bm{\tilde{q}}}_K\}),
%\label{eq:ins}
%\end{align}
\begin{align}
\footnotesize{\textit{instance selection:}}~~~& \bm{Q}={f}_\text{sel}(\bm{B}^{\prime}_\text{F})\in\mathbb{R}^{K\times{C}}, \\
\footnotesize{\textit{context aggregation:}}~~~&\tilde{\bm{{Q}}}={f}_\text{agg}(\bm{Q}, \bm{B}^{\prime}_\text{F})\in\mathbb{R}^{K\times{C}}, \\
\footnotesize{\textit{instance-to-scene:}}~~~& \hat{\bm{{B}}}_\text{F}={f}_\text{I2S}(\bm{B}^{\prime}_\text{F}, \tilde{\bm{{Q}}})\in\mathbb{R}^{W\times{H}\times C},
%\vspace{-4mm}
\label{eq:ins}
\end{align}
where $f_\text{sel}(\cdot)$ selects the top-$K$ salient instance features $\bm{Q}\!=\![\bm{q}_\text{ins}^{1}, \bm{q}_\text{ins}^{2}, \ldots, \bm{q}_\text{ins}^K]$, ${f}_\text{agg}(\cdot)$ aggregates the multimodal context for each instance and ${f}_\text{I2S}(\cdot)$ merges the augmented instance features $\tilde{\bm{{Q}}}\!=\![\tilde{\bm{{q}}}_\text{ins}^{1}, \tilde{\bm{{q}}}_\text{ins}^{2}, \ldots, \tilde{\bm{{q}}}_\text{ins}^K]$ to the BEV scene feature $ \bm{{B}}^{\prime}_\text{F}$. We present the overall pipeline of IGF in Fig.~\ref{fig:IGF}, and explain ${f}_\text{sel}(\cdot)$,  ${f}_\text{agg}(\cdot)$ and ${f}_\text{I2S}(\cdot)$ as follows.

%In a typical outdoor scene represented by BEV (as seen in Fig.~\ref{fig:heatmap}), the object instances often exhibit considerably small sizes. For example, a pedestrian or a traffic cone may occupy just one or two grid cells in a 180$\times$180 BEV feature map (\eg, covering an area around $100 \times 100\,\text{m}^2$ in nuScene). \textcolor{red}{Additionally, the spatial distribution of the objects in BEV is quite sparse due to the expanded perception area compared to the perspective-view images}, posing challenges for accurately recognizing outdoor objects. Motivated by these observations, we propose the Instance-Guided Fusion (IGF) module to emphasize fusion on the informative grid cells potentially containing objects instances instead of the background ones. 
%Since the BEV grids containing object instances are more informative, it is required to emphasize them more than other background grids. 
%To this end, we present the Instance-Guided Fusion (IGF) module that further delves into the instance-level multimodal encoding.
% The basic idea of IGF is to mine the multimodal context relevant to the instances (\eg, the lanes besides the vehicles) and then passes essential information to the scene-level feature to guide the fusion process. This leads to an instance-aware BEV representation that is crucial for tasks like 3D object detection. As illustrated in Fig.~\ref{fig:IGF}, IGF consists of three necessary components: instance candidates generation, instance context aggregation and an Instance-to-Scene transformer.

\noindent\textbf{Instance Candidates Selection.} To efficiently generate instance features, we implement ${f}_\text{sel}(\cdot)$ following~\cite{yin2021center} that applies a keypoint detection head on the scene feature $\bm{B}^{\prime}_\text{F}$ to predict the centerness of instances. During training, a 2D Gaussian distribution is defined for each instance as the target, and the peak location is determined by the BEV projection of the 3D center of the ground truth. Focal loss is employed to optimize this prediction head.
%enforce $\bm{H}$ to estimate the probability of object centers. 
During inference, we keep the top-$K$ object with the highest centerness scores to represent the corresponding instances. Meanwhile, an additional linear layer is employed to embed each instance, yielding a set of instance features $\{\bm{q}_\text{ins}^{1}, \bm{q}_\text{ins}^{2}, \ldots, \bm{q}_\text{ins}^K\}$.

%----------- Figure 4 ----------- %

\begin{table*}
\centering
\renewcommand\arraystretch{1.00}
\resizebox{0.98\textwidth}{!}
{
\begin{tabular}{r@{}l|c|cc|ccccccccccc}
\hline\thickhline
\multicolumn{2}{r|}{\textbf{Method}}~~~~~~~~~~~~~~ &   \textbf{Modality}  &  \textbf{mAP}  &  \textbf{NDS}  & \textbf{Car} & \textbf{Truck} &  \textbf{C.V.} &  \textbf{Bus} &  \textbf{T.L.} &  \textbf{B.R.} & \textbf {M.T.} &  \textbf{Bike} &  \textbf{Ped.} &  \textbf{T.C.} \\ \hline \hline
      CenterPoint~\cite{yin2021center}&~\textcolor[HTML]{708090}{\scriptsize{[CVPR 21]}}   &L     & 58.0               & 65.5       & 84.6         & 51.0           & 17.5          & 60.2         & 53.2             & 70.9             & 53.7           & 28.7          & 83.4          & 76.7          \\
      Focals Conv~\cite{chen2022focal}&~\textcolor[HTML]{708090}{\scriptsize{[CVPR 22]}}  &L      & 63.8             & 70.0             & 86.7         & 56.3           & 23.8          & 67.7         & 59.5             & 74.1             & 64.5           & 36.3          & 87.5          & 81.4          \\
       VoxelNeXt~\cite{chen2023voxelnext}&~\textcolor[HTML]{708090}{\scriptsize{[CVPR 23]}}     &L        & 64.5           & 70.0          & 84.6         & 53.0           & 28.7          & 64.7         & 55.8             & 74.6             & 73.2           & 45.7          & 85.8          & 79.0          \\
    TransFusion-L~\cite{bai2022transfusion}&~\textcolor[HTML]{708090}{\scriptsize{[CVPR 22]}}   &L         & 65.5       & 70.2               & 86.2         & 56.7           & 28.2          & 66.3         & 58.8             & 78.2             & 68.3           & 44.2          & 86.1          & 82.0          \\ 
      PillarNet-34~\cite{shi2022pillarnet}&~\textcolor[HTML]{708090}{\scriptsize{[ECCV 22]}}  &L    & 66.0              & 71.4               & 87.6         & 57.5           & 27.9          &63.6         & 63.1             & 77.2             & 70.1           & 42.3          & 87.3         & 83.3          \\ 
%LiDARMultiNet~\cite{ye2023lidarmultinet}     & L                 & 71.6         & 67.0         & 86.9         & 57.4           & 31.5         &64.7         & 61.0             & 73.5             & 75.3           & 47.6         & 87.2          & 85.1         \\ 
      FocalFormer3D~\cite{chen2023focalformer3d}&~\textcolor[HTML]{708090}{\scriptsize{[ICCV 23]}} &L    & 68.7               & 72.6             & 87.2         & 57.1           & 34.4          & 69.6         & 64.9             & 77.8             & 76.2           & 49.6        & 88.2         & 82.3         \\ 

\hline \hline
             MVP~\cite{yin2021multimodal}&~\textcolor[HTML]{708090}{\scriptsize{[NeurIPS 21]}}  &L+C   & 66.4             & 70.5           & 86.8         & 58.5           & 26.1          & 67.4         & 57.3             & 74.8             & 70.0           & 49.3          & 89.1          & 85.0          \\
            GraphAlign~\cite{song2023graphalign}&~\textcolor[HTML]{708090}{\scriptsize{[ICCV 23]}} &L+C      & 66.5             & 70.6           & 87.6         & 57.7           & 26.1          & 66.2         & 57.8             & 74.1             & 72.5           & 49.0          & 87.2          & 86.3          \\
     PointAug.~\cite{wang2021pointaugmenting}&~\textcolor[HTML]{708090}{\scriptsize{[CVPR 21]}}  &L+C       & 66.8       & 71.1                 & 87.5         & 57.3           & 28.0          & 65.2         & 60.7             & 72.6             & 74.3           & 50.9          & 87.9          & 83.6          \\
           UVTR~\cite{li2022unifying}&~\textcolor[HTML]{708090}{\scriptsize{[NeurIPS 22]}}  &L+C   & 67.1            & 71.1          &  87.5         &       56.0      &     33.8       &    67.5    &   59.5   &    73.0   &  73.4   &    54.8 &    86.3    &  79.6 \\
%  FusionPainting~\cite{xu2021fusionpainting} &  &L+C     & 68.1           & 71.6                & 87.1         & 60.8           & 30.0          & 68.5         & 61.7             & 71.8             & 74.7           & 53.5          & 88.3          & 85.0          \\
     AutoAlignV2~\cite{chen2022autoalignv2}&~\textcolor[HTML]{708090}{\scriptsize{[ECCV 22]}}  &L+C       & 68.4         & 72.4              & 87.0         & 59.0           & 33.1          & 69.3         & 59.3             & -                & 72.9           & 52.1          & 87.6          & -             \\
TransFusion-LC~\cite{bai2022transfusion}&~\textcolor[HTML]{708090}{\scriptsize{[CVPR 22]}}  &L+C        & 68.9        & 71.7               & 87.1         & 60.0           & 33.1          & 68.3         & 60.8             & 78.1             & 73.6           & 52.9          & 88.4          & 86.7          \\
       BEVFusion~\cite{liang2022bevfusion}&~\textcolor[HTML]{708090}{\scriptsize{[NeurIPS 22]}}   &L+C       & 69.2         & 71.8             & 88.1         & 60.9           & 34.4          & 69.3         & 62.1             & 78.2             & 72.2           & 52.2          & 89.2          & 85.5          \\
     BEVFusion~\cite{liu2022bevfusion}&~\textcolor[HTML]{708090}{\scriptsize{[ICRA 23]}}  &L+C       & 70.2           & 72.9             & 88.6         & 60.1           & 39.3          & 69.8         & 63.8             & 80.0             & 74.1           & 51.0          & 89.2          & 86.5          \\
 DeepInteraction~\cite{yang2022deepinteraction}&~\textcolor[HTML]{708090}{\scriptsize{[NeurIPS 22]}}  &L+C       & 70.8        & 73.4               & 87.9         & 60.2           & 37.5          & 70.8         & 63.8             & 80.4             & 75.4           & 54.5          & 91.7          & 87.2          \\   
 UniTR~\cite{wang2023unitr}&~\textcolor[HTML]{708090}{\scriptsize{[ICCV 23]}}  &L+C   & 70.9  & 74.5              &  87.9            &    60.2            &      39.2       &     72.2       &        65.1       &      76.8          &  75.8            &  52.2         &     89.4      &     89.7    \\
ObjectFusion~\cite{cai2023objectfusion}&~\textcolor[HTML]{708090}{\scriptsize{[ICCV 23]}}  &L+C    & 71.0  & 73.3              &    89.4          &      59.0          &      40.5         &      71.8        &    63.1              &       80.0           &      78.1          &  53.2             &      90.7         &   87.7        \\
MSMDFusion~\cite{cai2023objectfusion}&~\textcolor[HTML]{708090}{\scriptsize{[CVPR 23]}}  &L+C    & 71.5  & 74.0              &    88.4          &      61.0          &      35.2         &      71.4        &    64.2              &       80.7           &      76.9          &  58.3             &      90.6         &   88.1        \\
    FocalFormer3D~\cite{chen2023focalformer3d}&~\textcolor[HTML]{708090}{\scriptsize{[ICCV 23]}}  &L+C     & 71.6           & 73.9               & 88.5         & 61.4           & 35.9          & 71.7         & 66.4             & 79.3             & 80.3           & 57.1          & 89.7          & 85.3          \\
  SparseFusion~\cite{li2023fully}&~\textcolor[HTML]{708090}{\scriptsize{[ICCV 23]}} &L+C  & 72.0              & 73.8                 & 88.0         & 60.2           & 38.7          & 72.0         & 64.9             & 79.2             & 78.5           & 59.8          & 90.9          & 87.9          \\
          CMT~\cite{yan2023cross}&~\textcolor[HTML]{708090}{\scriptsize{[ICCV 23]}}   &L+C     & 72.0              & 74.1           & 88.0         & 63.3           & 37.3          & 75.4         & 65.4             & 78.2             & 79.1           & 60.6          & 87.9          & 84.7        
    \\\hline
\multicolumn{2}{r|}{\ourmodel~(Ours)}~~    &L+C    & {73.0}          & {75.2}               & 88.3             & 62.7               & {38.4}              &   74.9           &     {67.3}             &       78.1           &    82.4          &    59.5             &   89.3           &    89.2            \\
\multicolumn{2}{r|}{\ourmodel$^{\dag}$~(Ours)}~~    &L+C    & \textbf{76.5}          & \textbf{77.4}               & \textbf{89.8}             & \textbf{67.8}               & \textbf{44.5}              &   \textbf{77.6}           &     \textbf{68.3}             &       \textbf{81.8}           &    \textbf{85.3}          &    \textbf{65.6}             &   \textbf{93.4}           &    \textbf{91.1}            \\
% Ours (\ourmodel$^\dag$)      & -  &L+C     &              &              &              &                &               &              &                 &                  &                &               &               &      \\
\thickhline 
        
\end{tabular}
}
%\caption{{\textbf{3D Object Detection Performance on the nuScenes test set.} `L' is the LiDAR sensor and `C' is the camera sensor. `C.V.', `T.L.', `B.R.', `M.T.', `Ped.', and ‘T.C.’ indicates the construction vehicle, trailer, barrier, motor, pedestrian and traffic cone, respectively. \ourmodel~is the single-model result. \ourmodel$^\dag$  denotes the result achieved by using test-time augmentation and model ensemble strategies.}}
% \vspace{-2mm}
\caption{{\textbf{3D Object Detection Performance on the nuScenes test set.} `L' is the LiDAR and `C' denotes the camera. `C.V.', `T.L.', `B.R.', `M.T.', `Ped.', and ‘T.C.’ indicate the construction vehicle, trailer, barrier, motorcycle, pedestrian, and traffic cone, respectively. `$\dag$' denotes the model with test-time augmentation and model ensemble techniques. The best results in each column are marked in bold font. \ourmodel~achieves superior performance compared to all the other published 3D detection works.}}
\label{tb:test}
% \vspace{-2mm}
\end{table*}

\noindent\textbf{Instance Context Aggregation.} We design ${f}_\text{agg}(\cdot)$ to count for both the instance-to-instance and instance-to-context interactions. In typical driving scenarios, it is often observed that pedestrians tend to appear in groups or clusters, and vehicles commonly co-exist along the roadside. {Thus, it is crucial to investigate the correlations between instances. To this end, we employ self-attention ${f}_\text{MSA}(\cdot)$ on the selected instances $[\bm{q}_\text{ins}^{1}, \bm{q}_\text{ins}^{2}, \ldots, \bm{q}_\text{ins}^K]\!\in\!\mathbb{R}^{K\times{C}}$:} 
%Specifically, let $\bm{h}\in\mathbb{R}^{K\times{C}}$ represent a group of instance $\{\bm{h}_0, \bm{h}_1, \ldots, \bm{h}_K\}$. A multi-head attention is first applied on $\bm{h}$:
%\begin{equation}
%\begin{aligned}
%&	 \bm{h}' = \text{LN}(\text{MSA}(\bm{h})+\bm{h}), \\
%&	 \hat{\bm{h}} = \text{LN}(\text{MLP}(\bm{h^{'}})+\bm{h}),
%\end{aligned}
%\end{equation}
\begin{equation}
%\begin{aligned}
[\bm{q}^{\prime1}_{\text{ins}}, \bm{q}^{\prime2}_{\text{ins}}, \ldots, \bm{q}^{\prime K}_{\text{ins}}] = {f}_\text{MSA}([\bm{q}_{\text{ins}}^{1}, \bm{q}_{\text{ins}}^{2}, \ldots, \bm{q}_{\text{ins}}^K]), 
%& \bm{g}_{I} = \max_{l}(\bm{p}'_{l}) \in \mathbb{R}^{1 \times C},
%\end{aligned}
\end{equation}
%where $\{\bm{q'}_0, \bm{q'}_1, \ldots, \bm{q'}_K\}$ is the attentive instance features. 
Furthermore, we aim to mine the semantic context for each instance. This is achieved by comparing each instance $\bm{q}^{\prime}_\text{ins}$ and the corresponding part in the multimodal feature $\bm{B}^{\prime}_\text{F}$. Specifically, only on a small set of neighbor locations (\eg, $D$ grid cells) around $\bm{q}^{\prime}_\text{ins}$ are considered to save computation costs, following the deformable attention in~\cite{zhu2020deformable}:
\begin{equation}
%\begin{aligned}
\tilde{\bm{{q}}}_\text{ins}= {f}_\text{DeformAtt}(\bm{q}^{\prime}_\text{ins}, f_\text{conv}(\bm{B}^{\prime}_\text{F})), 
%& \bm{g}_{I} = \max_{l}(\bm{p}'_{l}) \in \mathbb{R}^{1 \times C},
%\end{aligned}
\end{equation}
where $f_\text{conv}(\cdot)$ is a $3\!\times\!{3}$ convolution operation to align the feature space between $\bm{q}^{\prime}_\text{ins}$ and $\bm{B}^{\prime}_\text{F}$, and $\tilde{\bm{{q}}}_\text{ins}\!\in\!\{\tilde{\bm{{q}}}^1_\text{ins},\tilde{\bm{{q}}}^2_\text{ins},$ $  \ldots, \tilde{\bm{{q}}}^K_\text{ins}\}$ is the enriched instance features.

\noindent\textbf{Instance-to-Scene Transformer.} Finally, ${f}_\text{I2S}(\cdot)$ enables each BEV grid feature to acquire valuable information from potentially relevant instances.
To this end, we build ${f}_\text{I2S}(\cdot)$ with a transformer cross-attention mechanism.
Specifically, after flattening $\bm{B}^{\prime}_\text{F}\!\in\!\mathbb{R}^{W\times{H}\times{C}}$ into a set of grid features $\{\bm{g}^{\prime1}_\text{grid}, \bm{g}^{\prime2}_\text{grid}, \ldots, \bm{g}^{\prime W\times{H}}_\text{grid}\}$, we employ each grid $\bm{g}^{\prime}_\text{grid}$ as a query to attend to the instance-level features $[\tilde{\bm{{q}}}^1_\text{ins}, \tilde{\bm{{q}}}^2_\text{ins}, \ldots,$ $ \tilde{\bm{{q}}}^K_\text{ins}]\in\mathbb{R}^{K\times{C}}$:	
%and each grid feature $\bm{g}$ serves as a query. Meanwhile, we exploit the group of instance features $\tilde{\bm{h}}=\{\tilde{\bm{h}}_0, \tilde{\bm{h}}_1, \ldots, \tilde{\bm{h}}_K\}$ as keys and values. Then, the instance-to-scene integration can be expressed as:
%\begin{equation}
%\begin{aligned}
%&	 \bm{x}' = \text{LN}(\text{MCA}({\bm{x}},  \tilde{\bm{h}})+\bm{x}), \\
%&	 \hat{\bm{x}} = \text{LN}(\text{MLP}(\bm{x^{'}})+\bm{x}),
%\end{aligned}
%\end{equation}
\vspace{-3mm}
\begin{equation}
%\begin{aligned}
{\hat{\bm{{g}}}_\text{grid}} = {f}_\text{MCA}(\bm{g}^{\prime}_\text{grid}, [\tilde{\bm{{q}}}^1_\text{ins}, \tilde{\bm{{q}}}^1_\text{ins}, \ldots, \tilde{\bm{{q}}}^K_\text{ins}]), 
%& \bm{g}_{I} = \max_{l}(\bm{p}'_{l}) \in \mathbb{R}^{1 \times C},
%\end{aligned}
\end{equation}
where ${f}_\text{MCA}(\cdot)$ indicates the multi-head cross-attention and $\hat{\bm{{g}}}_\text{grid}$ is an attentive grid cell. After applying ${f}_\text{MCA}(\cdot)$ on all the grid cells, we rearrange the obtained grid features $\{\hat{\bm{{g}}}^1_\text{grid}, \hat{\bm{{g}}}^2_\text{grid}, \ldots, \hat{\bm{{g}}}^{W\times{H}}_\text{grid}\}$ back into a BEV feature ${\hat{\bm{{B}}}_\text{F}}\!\in\!\mathbb{R}^{W\times{H}\times{C}}$, which will be employed in the subsequent decoding stage to produce the final 3D detections. 
%We will demonstrate in Sec.~\ref{sec:ablation} that IGF enables the BEV scene feature to capture more meaningful context around instances, leading to improved representation quality.

%----------- Figure 5 ----------- %
\begin{figure*}[t]
\includegraphics[width=0.95\linewidth]{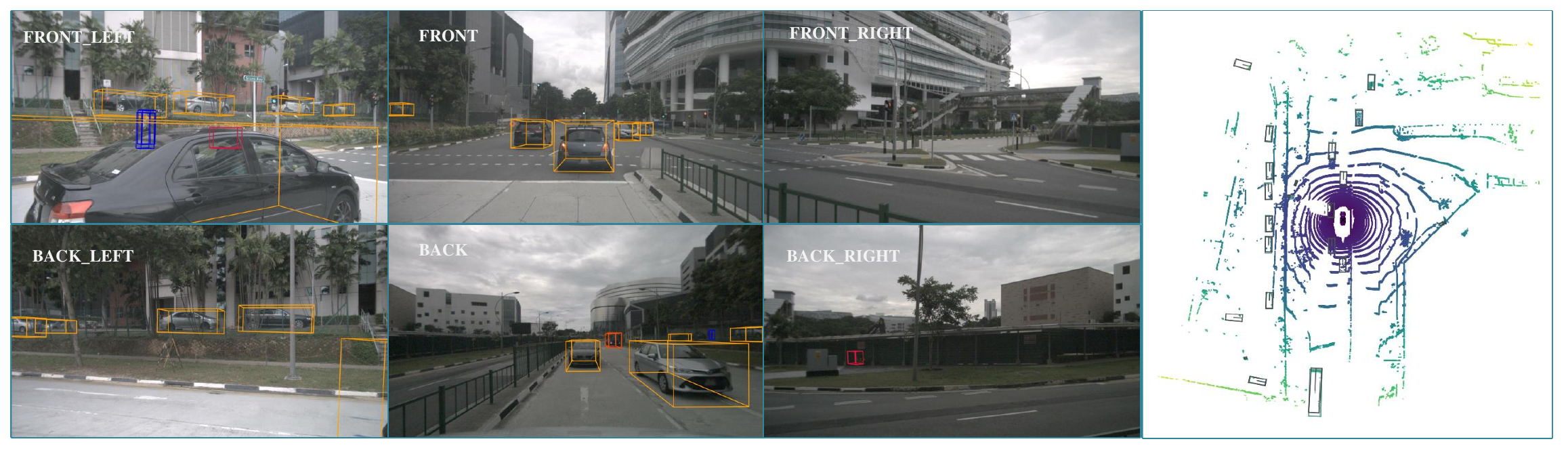}
\centering
 % \vspace{-3mm}
 \caption{ \textbf{Examples of 3D object detections} on nuScenes validation set. We visualize the 3D bounding boxes of car, pedestrian and bicycle with \textbf{{\textcolor{orange}{orange}}},  \textbf{{\color{blue!80}{blue}}} and \textbf{{\color{red!90}{red}}} colors in the multi-view images. In the point cloud, the predictions are in  \textbf{{\color{gray}{gray}}} and GTs are in  \textbf{{\textcolor{teal}{green}}}.}
\label{fig:detection}
% \vspace{-4mm}
\end{figure*}

\section{Experiments} 
\label{experiments}

\subsection{Experimental Setup} 
\noindent\textbf{Dataset.} We evaluate the 3D object detection performance of the proposed \ourmodel~by comparing it with other state-of-the-art approaches on the nuScenes benchmark~\cite{caesar2020nuscenes}. nuScenes is a very challenging large-scale autonomous driving dataset that is widely used for evaluating multi-modality 3D object detectors. It provides 700, 150, and 150 scene sequences for training, validation, and testing, respectively. Each sequence in the dataset consists of approximately 40 frames of annotated LiDAR point cloud data, and each point cloud data is accompanied by six calibrated image data covering $360^\circ$ field of view. It requires detecting 10 object categories that are commonly observed in driving scenarios. The evaluation of 3D object detection is based on two key metrics: mean Average Precision (mAP) and nuScenes detection scores (NDS). In particular, NDS is a comprehensive metric that consolidates object translation, scale, orientation, velocity and attribute.

%\vspace{-2mm}
\noindent\textbf{Network Architecture.} Our implementation follows the open-source framework MMDetection3D~\cite{mmdet3d2020}. Specifically, the point cloud covers [-54$m$, 54$m$] along the X and Y axes, and [-5$m$, 3$m$] along the Z axis, with a voxel size of (0.075$m$, 0.075$m$, 0.2$m$). In the Point-to-Grid transformer, we set the pillar size to (0.6$m$, 0.6$m$, 8.0$m$). The input resolution of multi-view images is set to $384\times1056$. The BEV feature map is of size $180\times180$. In HSF, we define the point number $L$ as 20 and the region size $M$ as 6. In IGF, the number of instance candidates $K$ is set to 200. The number of sampling locations $D$ on the multimodal feature is set to 16. For the model ensemble, multiple models are utilized with voxel sizes ranging from (0.05m, 0.05m, 0.2m) to (0.125m, 0.125m, 0.2m) with intervals of 0.025m. For the test-time augmentation, we apply double flipping and rotations (\ie, \{$0\degree$, $\pm22.5\degree$,  $\pm180\degree$\}) on the input point clouds. 

\noindent\textbf{Training.} The image encoder is pre-trained on the nuImage dataset~\cite{caesar2020nuscenes} following current approaches~\cite{liu2022bevfusion,yang2022deepinteraction,bai2022transfusion}. The full model is trained end-to-end  for 10 epochs with the AdamW optimizer~\cite{loshchilov2017decoupled}. Meanwhile, the once-cycle learning policy~\cite{smith2017cyclical} is employed with a maximum learning rate of $1e^{-3}$. The class-balanced sampling strategy from CBGS~\cite{zhu2019class} and the cross-modal data augmentation from AutoAlignV2~\cite{chen2022autoalignv2} are adopted during training. The design of the 3D decoder follows the common practices of leading approaches, such as TransFusion-L~\cite{bai2022transfusion} and BEVFusion~\cite{liu2022bevfusion}, where we decode the top 200 bounding boxes.

\begin{table}[t]
\renewcommand\arraystretch{1.00}
\resizebox{0.48\textwidth}{!}
{
\begin{tabular}{r@{}l|l|ll|l}
\hline\thickhline
\multicolumn{2}{r|}{{Method}}~~~~~~~~~~~~     & \begin{tabular}[c]{@{}l@{}}Image\\ Encoder\end{tabular} & mAP  & NDS & FPS \\ \hline\hline
FUTR3D~\cite{chen2023futr3d} & ~\textcolor[HTML]{708090}{\scriptsize{[CVPRW 23]}}                      & ResNet-101                     & 64.2                            & 68.0      & 2.3 \\
%MVP~\cite{yin2021multimodal}                             & DLA-34                                                   & 70.8     & 67.1   & 5.3 \\
TransFusion-LC~\cite{bai2022transfusion}     &~\textcolor[HTML]{708090}{\scriptsize{[CVPR 22]}}                & ResNet-50                          & 67.5                        & 71.3      & 3.2 \\
BEVFusion~\cite{liu2022bevfusion}  &~\textcolor[HTML]{708090}{\scriptsize{[ICRA 23]}}                     & Swin-T                      & 68.5                               &   71.4        & 4.2 \\
DeepInteraction~\cite{yang2022deepinteraction}&~\textcolor[HTML]{708090}{\scriptsize{[NeurIPS 22]}}                  & Swin-T                  & 69.9                                  & 72.6       & 2.6 \\
CMT~\cite{yan2023cross}   &~\textcolor[HTML]{708090}{\scriptsize{[ICCV 23]}}                          & VoV-99                                    & 70.3                 & 72.9      & 3.8 \\
%UniTR~\cite{wang2023unitr}         & ViT                                    & 70.5                 & 73.3      & \textbf{9.3} \\
SparseFusion~\cite{li2023fully} &~\textcolor[HTML]{708090}{\scriptsize{[ICCV 23]}}                   & Swin-T                              & 71.0                         & 73.1    & \textbf{5.3} \\ \hline
\multicolumn{2}{r|}{{\ourmodel~(Ours)}}~~ & Swin-T                           & \textbf{72.8}                          &  \textbf{74.0}      & 3.2 \\ \thickhline
\end{tabular}
}
\caption{{\textbf{Performance comparison on the nuScenes validation set.} \ourmodel~achieves superior 3D detection performance while maintaining a comparable inference speed.}}
\label{tb:val}
% \vspace{-4mm}
\end{table}

% \vspace{-2mm}
\subsection{Performance Benchmarking}
In Table~\ref{tb:test}, we benchmark the performance of our model against current leading LiDAR-based (indicated by `L') and multimodal (indicated by `L+C') 3D object detectors on the nuScenes test set. It demonstrates that \ourmodel~outperforms all existing state-of-the-art (SOTA) 3D detection algorithms. Specifically, the LiDAR-only baseline of \ourmodel~is built upon TransFusion-L~\cite{bai2022transfusion}. By exploring instance-scene collaborative fusion, \ourmodel~significantly improves it by 7.5\% in mAP and 5.0\% in NDS, respectively. Furthermore, \ourmodel~demonstrates superior performance compared to some very recent multimodal detection works such as FocalFormer3D~\cite{chen2023focalformer3d}, SparseFusion~\cite{li2023fully} and CMT~\cite{yan2023cross}, outperforming them by 1.4\%, 1.0\% and 1.0\% in mAP, respectively. Notably, \ourmodel~obtains the highest results in some categories with fewer labeled instances, \ie, motorcycle and trailers (constituting only 1.08\% and 2.13\% of the dataset). This suggests that \ourmodel~captures essential information even from limited instances. By applying test-time augmentation and model ensemble, $\ourmodel^{\dag}$~achieves a new SOTA on the highly competitive nuScenes leaderboard.

As shown in Table~\ref{tb:val}, \ourmodel~also obtains the best detection accuracy on the nuScenes validation set, meanwhile keeping a comparable inference speed. In particular, it significantly surpasses the SOTA detectors like CMT and SparseFusion by 2.5\% and 1.8\% in mAP, respectively. In Fig.~\ref{fig:detection}, we additionally present some qualitative detection results on the nuScenes validation set to showcase the performance of \ourmodel. The visualization reveals that \ourmodel~is capable of accurately detecting objects of various classes, even at distant ranges and with varying scales. Overall, the promising performance of \ourmodel~can be attributed to the joint modeling of the multimodal instance-level and scene-level contexts, as well as their effective collaboration in enhancing the BEV representation.

%This enhanced multimodal representation that is more suitable for describing complex driving scenarios, and thus benefits the 3D object detection.

%----------- Table 3 ----------- %
\begin{table}
\centering
\renewcommand\arraystretch{1.00}
\resizebox{0.48\textwidth}{!}
{
\begin{tabular}{l|cccc|ll}
\hline\thickhline
    & Baseline-L & Baseline-LC & HSF & IGF  & mAP & NDS \\ \hline \hline
(a) & \checkmark
          &             &     &   &65.4  &70.1          \\
(b) &           &  \checkmark           &     &   & 69.4  & 71.6        \\  \hline \hline
(c) &           &  \checkmark          &  \checkmark   &   & 71.6  &73.2          \\
(d) &           &  \checkmark          &     &  \checkmark & 70.9     &72.8       \\ 
(e) &           &  \checkmark           &  \checkmark   &  \checkmark &\textbf{72.8}  &\textbf{74.0}          \\ \thickhline
\end{tabular}
}
\caption{{\textbf{Ablation studies for each module in \ourmodel}~on the nuScenes validation set. Baseline-L indicates the LiDAR-only baseline, while Baseline-LC refers to a simple variant of  \ourmodel~without employing the HSF or IGF modules.}}
\label{tb:ablation}
\vspace{-4mm}
\end{table}

%----------- Table 4 ----------- %
\begin{table*}
     \begin{minipage}[t]{0.32\textwidth}
    \centering
    \resizebox{1.1\textwidth}{!}
{
%   \begin{tabular}{l|ll} \hline \thickhline
%        Paradigms          & NDS & mAP \\ \hline  \hline
%% Cross-Attn. &  71.3   &  68.1   \\
%Depth Estimation  &  71.3   &  68.4   \\
%Reference Points    & 71.5    &   68.2  \\ 
%Ours (Baseline-LC) &  71.6   & 69.4  \\ \thickhline
%\end{tabular}}
%    \subcaption{Different solutions for getting multimodal BEV feature in Baseline-LC.}
%    \label{tb:bev}
%  \end{minipage}\hfill
   \begin{tabular}{r|l|ll} \hline \thickhline
        Image Encoder & Resolution         &mAP  & NDS \\ \hline  \hline
% Cross-Attn. &  71.3   &  68.1   \\
ResNet-50~\cite{he2016deep}  & $320\times800$ & 71.3   &  72.8   \\
CSPNet~\cite{wang2020cspnet}    &  $384\times1056$ & 71.7    &   73.1  \\ 
Swin-T~\cite{liu2021swin} & $256\times704$ & 72.4   & 73.7  \\ 
Swin-T~\cite{liu2021swin} & $384\times1056$ & \textbf{72.8}   & \textbf{74.0}  \\\thickhline
\end{tabular}}
    \subcaption{Performance with different image encoders}
    \label{tb:img}
  \end{minipage}\hfil ~~~~~~
  \begin{minipage}[t]{0.32\textwidth}
    \centering
     \resizebox{0.9\textwidth}{!}
{
    \begin{tabular}{r|ll} \hline \thickhline
        Components          & mAP & NDS \\ \hline  \hline
        Baseline-LC  & 69.4    & 71.6    \\   \hline
+ Point-to-Grid Attn.  &  69.9   & 71.9   \\
+ Grid-to-Region Attn. & 71.2    &72.8   \\  
Full HSF module & \textbf{71.6}    &  \textbf{73.2}    \\ \thickhline
\end{tabular}}
    \subcaption{Component-wise ablation studies on HSF}
    \label{tb:hsf}
  \end{minipage}
\hspace{-3mm}  \begin{minipage}[t]{0.32\textwidth}
    \centering
         \resizebox{0.81\textwidth}{!}
{
    \begin{tabular}{r|ll} \hline \thickhline
    Hyper-parameter              & mAP &NDS  \\ \hline  \hline
$K$=64; $D$=16  & 69.8    &72.0     \\
$K$=200; $D$=16   & \textbf{70.9}    & \textbf{72.8}    \\
$K$=200; $D$=32   & 70.6    & 72.7    \\
$K$=300; $D$=16   & 70.4    &72.5     \\    \thickhline
\end{tabular}}
%     \resizebox{0.95\textwidth}{!}
%{
%  \begin{tabular}{l|ll} \hline
%Image Backbones                 & NDS & mAP \\ \hline
%ResNet-50  &     &     \\
%ResNet-101    &     &     \\
%Swin-Transformer &     &   \\  \hline
%\end{tabular}}
   \subcaption{Number of instances and neighbors in IGF}
   \label{tb:igf}
  \end{minipage}
    \vspace{-2mm}
\caption{{\textbf{Design choices of~\ourmodel.} We explore the impact of various components in HSF and the optimal hyper-parameters in IGF.}}
  \label{tb:design}
  \vspace{-4mm}
\end{table*}

\subsection{Ablation Studies}
\label{sec:ablation}

\subsubsection{Component-wise Ablation}
\label{sec:ablation}
In this section, we investigate the contribution of each component in our model. We begin by introducing the baseline frameworks of  \ourmodel. Concretely, our LiDAR-only baseline derives from Transfusion-L~\cite{bai2022transfusion}, which is reimplemented here as Baseline-L. For the multimodal baseline, denoted as {Baseline-LC}, we adopt a straightforward approach that combines the point cloud and image BEV features via a convolutional layer (see Eq.~(\ref{eq:bevfusion})). To obtain the image BEV features, we summarize the point features in each pillar through summation operation, where the point features are determined by the image features as introduced in Eq.~(\ref{eq:proj}). According to Table~\ref{tb:ablation}~{(a)}-{(b)}, this intuitive fusion solution achieves 69.4\% mAP and 71.6\% NDS, outperforming Baseline-L by 4.0\% in mAP and 1.5\% in NDS. By leveraging HSF to enhance the scene feature, it improves 2.2\% in mAP and 1.6\% in NDS in terms of Table~\ref{tb:ablation}~{(b)}-{(c)}. To verify the effect of instance-level modeling (\ie, Table~\ref{tb:ablation}~{(b)}-{(d)}), it shows that IGF outperforms Baseline-LC by 1.5\% in mAP and 1.2\% in NDS. This highlights the crucial role of instance representation. In Table~\ref{tb:ablation}~{(e)}, our full model, utilizing both HSF and IGF, achieves the best performance of 72.8\% mAP and 74.0\% NDS, demonstrating the effect of instance-scene collaboration.

Additionally, in Table \ref{tb:design}\subref{tb:img}, we explore the impact of different image encoders and input resolutions.  It shows that Swin-T~\cite{liu2021swin} outperforms other image encoders, such as ResNet-50~\cite{he2016deep} and CSPNet~\cite{wang2020cspnet}. It indicates that utilizing more powerful image encoders can potentially enhance the detection performance of \ourmodel. Furthermore, using a larger input image resolution (e.g., $384\times1056$) also leads to a slight performance improvement.

%----------- Figure 6 ----------- %
\begin{figure}
\includegraphics[width=0.95\linewidth]{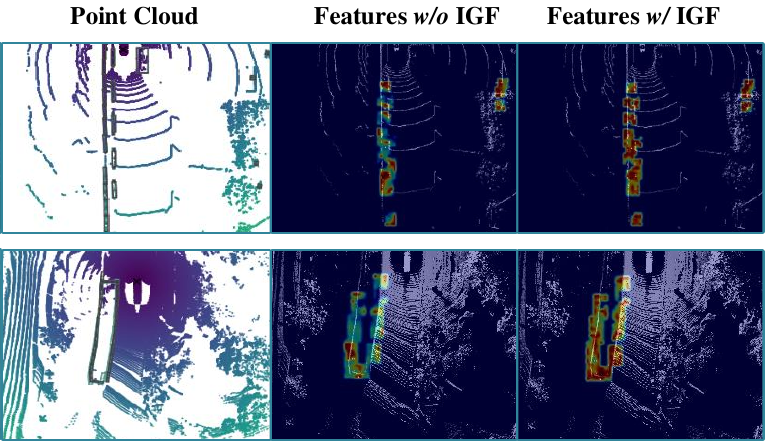}
\centering
 % \vspace{-2mm}
 \caption{ \textbf{Visualization of the BEV features} in challenging scenarios with traffic cones. We show the BEV features of models \textit{w/} and \textit{w/o} IGF. It demonstrates that IGF can yield instance representation with higher responses and more complete patterns.}
\label{fig:heatmap}
 % \vspace{-4mm}
\end{figure}

\vspace{-2mm}
\subsubsection{Analysis of HSF}
\vspace{-1mm}
%The HSF module is designed to hierarchically integrate image features from various granularities, enabling a comprehensive description of the scene context. Since we need to leverage the multimodal BEV feature, we first compare our Baseline-LC with other existing solutions for obtaining the image BEV feature. For example, BEVFusion~\cite{liu2022bevfusion} applies an extra module for estimating the depths of image pixels and then transforms the pseudo points to BEV, while UVTR~\cite{li2022unifying} exploits uniformly sampled reference points to fetch image features. As shown in Table \ref{tb:design}\subref{tb:bev}, our intuitive solution, Baseline-LC, outperforms these solutions by 0.6\% and 0.4\% in terms of NDS. Furthermore, we examine the effectiveness of the multi-granularity features in HSF. According to Table \ref{tb:design}\subref{tb:hsf}, both the Point-to-Grid and Grid-to-Region transformer attentions are crucial for the detection performance, which improve Baseline-LC by 0.5\% mAP and 0.9\% mAP, respectively. The full HSF module further improves the baseline by 0.9\% NDS and 1.6\% mAP. Moreover, the HSF also improves the quality of instance features, which enhances the IGF module by 0.6\% NDS and 1.0\% mAP, as indicated by \textbf{d)}-\textbf{e)} in Table~\ref{tb:ablation}.
The HSF module is designed to hierarchically extract multimodal features at various granularities, facilitating a comprehensive description of the scene context. Therefore, we examine the effectiveness of using different feature granularities in HSF. According to Table \ref{tb:design}\subref{tb:hsf}, the Point-to-Grid transformer, focusing on point-wise and grid-wise features, shows an improvement over Baseline-LC by 0.5\% in mAP and 0.3\% in NDS. The Grid-to-Region transformer improves Baseline-LC by 1.8\% in mAP and 1.2\% in NDS, by exploring the inter-grid and inter-region features. It suggests that a larger receptive field is more crucial for 3D object detection. The full HSF results in an improvement of 2.2\% in mAP and 1.6\% in NDS, highlighting the benefits of feature integration across different granularities.

\vspace{-1mm}
\subsubsection{Analysis of IGF}
\vspace{-1mm}
The IGF module aggregates the local multimodal feature around each instance, and incorporates necessary instance-level information into the BEV scene feature. In IGF, there are two hyper-parameters that need to be determined, namely, the instance number ($K$) and the sampled neighbor number ($D$) in the multimodal feature.  According to Table~\ref{tb:design}\subref{tb:igf}, we found that setting $K=200$ and $D=16$ yields better performance, achieving 70.9\% mAP and 72.8\% NDS. Further increasing $K$ or $D$ does not lead to additional improvement, which suggests that the self-attention between instances has effectively explored a suitable receptive field. Additionally, we provide visualization of the BEV feature maps for models with and without IGF. As shown in Fig.~\ref{fig:heatmap}, the feature maps without IGF tend to exhibit incomplete patterns and lower responses, while the IGF module significantly enhances the quality of the feature map, due to the interactive collaboration with the instance-level feature.

%\subsubsection{Influence of various image backbones.}

\section{Conclusions}

This work presents an innovative fusion framework, \ourmodel, for multimodal 3D object detection. It consists of two essential modules, \ie, the Hierarchical Scene Fusion (HSF) module and the Instance-Guided Fusion (IGF) module. 
In particular, Point-to-Grid and Grid-to-Region transformer attentions are designed in HSF to capture hierarchical scene context. Furthermore, IGF is introduced to mine instances, explore inter-instance relationships and incorporate rich multimodal context around instance. We also propose an Instance-to-Scene transformer attention to encourage the collaboration between the instance and scene representations. \ourmodel~achieves superior performance on the competitive nuScenes benchmark. It provides a fresh perspective to current BEV-based perception models by emphasizing instance-level context, which is potentially beneficial to a spectrum of instance-centric tasks.

%HSF aims to capture hierarchical scene context by incorporating multimodal features from different granularities. This is realized by using Point-to-Grid and Grid-to-Region transformer attentions. Furthermore, IGF focuses on exploring inter-instance relationships while incorporating rich multimodal context. An Instance-to-Scene transformer attention mechanism is also employed to integrate the instances into the scene to get an instance-aware BEV representation. \ourmodel~achieves top performance on the competitive nuScenes benchmark. 

% \noindent\textbf{Acknowledgements.} 
% This work was supported in part by the FDCT grants
% 0154/2022/A3, 
% 0102/2023/RIA2, and 001/2024/SKL, the MYRG-CRG2022-00013-IOTSC-ICI grant, the SRG2022-00023-IOTSC grant, the General Program of the National Natural Science Foundation of China (No. 62372405) and CCF-Tencent Open Fund.

%%%%%%%%% REFERENCES
{\small
\bibliographystyle{ieee_fullname}
\bibliography{InsFusion}
}

\end{document}